%% file: main.tex
\documentclass[10pt, conference, compsocconf]{IEEEtran}
\usepackage{cite}

\ifCLASSINFOpdf
  \usepackage[pdftex]{graphicx}
\else
\fi
\usepackage{amsmath,amsfonts,amsthm,mathtools,bm}

\usepackage[small]{caption}
\usepackage{url}

\usepackage{todonotes}

\usepackage{subfig}
\usepackage{tikz}
\usepackage{tikzscale}
\usepackage{pgfplots}
\usepackage[colorlinks,
urlcolor=blue,
citecolor=black!50!green,
linkcolor=blue
]{hyperref}
\usepackage{cleveref}

\usetikzlibrary{arrows,arrows.meta}
\usetikzlibrary{positioning,shapes,shapes.geometric,automata,fit,calc,decorations.pathreplacing,angles,quotes,backgrounds}
\usepgfplotslibrary{external}
\tikzset{%
	every loop/.style={in=50,out=130,looseness=3},
	every node/.style={thin, font=\small},
	>=stealth
}

\definecolor{tucol-kraeftig1}{RGB}{132,184,25}
\definecolor{tucol-kraeftig2}{RGB}{216,148,39}
\definecolor{tucol-kraeftig3}{RGB}{27,161,175}
\definecolor{tucol-kraeftig4}{RGB}{168,0,135}
\definecolor{tucol-kraeftig5}{RGB}{239,228,32}
\definecolor{tucol-kraeftig6}{RGB}{202,116,40}

\definecolor{tucol-kuehl1}{RGB}{132,184,25}
\definecolor{tucol-kuehl2}{RGB}{233,238,168}
\definecolor{tucol-kuehl3}{RGB}{217,233,229}
\definecolor{tucol-kuehl4}{RGB}{191,218,187}
\definecolor{tucol-kuehl5}{RGB}{113,176,96}
\definecolor{tucol-kuehl6}{RGB}{211,223,99}

\definecolor{tucol-warm1}{RGB}{132,184,25}
\definecolor{tucol-warm2}{RGB}{112,61,46}
\definecolor{tucol-warm3}{RGB}{193,178,40}
\definecolor{tucol-warm4}{RGB}{228,200,38}
\definecolor{tucol-warm5}{RGB}{178,175,132}
\definecolor{tucol-warm6}{RGB}{165,134,83}

\definecolor{tucol-rot1}{RGB}{132,184,25}
\definecolor{tucol-rot2}{RGB}{196,21,58}
\definecolor{tucol-rot3}{RGB}{157,19,42}
\definecolor{tucol-rot4}{RGB}{119,24,40}
\definecolor{tucol-rot5}{RGB}{97,126,31}
\definecolor{tucol-rot6}{RGB}{58,82,11}

\definecolor{tucol-gruen1}{RGB}{132,184,25}
\definecolor{tucol-gruen2}{RGB}{214,223,42}
\definecolor{tucol-gruen3}{RGB}{97,134,39}
\definecolor{tucol-gruen4}{RGB}{148,164,33}
\definecolor{tucol-gruen5}{RGB}{182,201,48}
\definecolor{tucol-gruen6}{RGB}{115,158,64}

\definecolor{tucol-blau1}{RGB}{132,184,25}
\definecolor{tucol-blau2}{RGB}{11,161,226}
\definecolor{tucol-blau3}{RGB}{36,123,196}
\definecolor{tucol-blau4}{RGB}{40,140,141}
\definecolor{tucol-blau5}{RGB}{177,213,230}
\definecolor{tucol-blau6}{RGB}{13,75,127}

\usepackage{xspace}
\input{macros}

\begin{document}
%
\title{Exploring Confidence Measures for Word Spotting in Heterogeneous Datasets}



\author{\IEEEauthorblockN{Fabian Wolf}
\IEEEauthorblockA{Department of Computer Science\\
TU Dortmund University\\
44227 Dortmund, Germany\\
fabian.wolf@cs.tu-dortmund.de}
\and
\IEEEauthorblockN{Philipp Oberdiek}
\IEEEauthorblockA{Department of Computer Science\\
TU Dortmund University\\
44227 Dortmund, Germany\\
philipp.oberdiek@cs.tu-dortmund.de}
\and
\IEEEauthorblockN{Gernot A. Fink}
\IEEEauthorblockA{Department of Computer Science\\
TU Dortmund University\\
44227 Dortmund, Germany\\
gernot.fink@cs.tu-dortmund.de}
}


%


\maketitle

\begin{abstract}
In recent years, convolutional neural networks (CNNs) took over the field of document analysis and they became the predominant model for word spotting.
Especially attribute CNNs, which learn the mapping between a word image and an attribute representation, showed exceptional performances.
The drawback of this approach is the overconfidence of neural networks when used out of their training distribution.
In this paper, we explore different metrics for quantifying the confidence of a CNN in its predictions, specifically on the retrieval problem of word spotting.
With these confidence measures, we limit the inability of a retrieval list to reject certain candidates.
We investigate four different approaches that are either based on the network's attribute estimations or make use of a surrogate model.
Our approach also aims at answering the question for which part of a dataset the retrieval system gives reliable results.
We further show that there exists a direct relation between the proposed confidence measures and the quality of an estimated attribute representation.
\end{abstract}

%

%

\section{Introduction}
\begin{figure*}[t]
	\includegraphics[width=\textwidth]{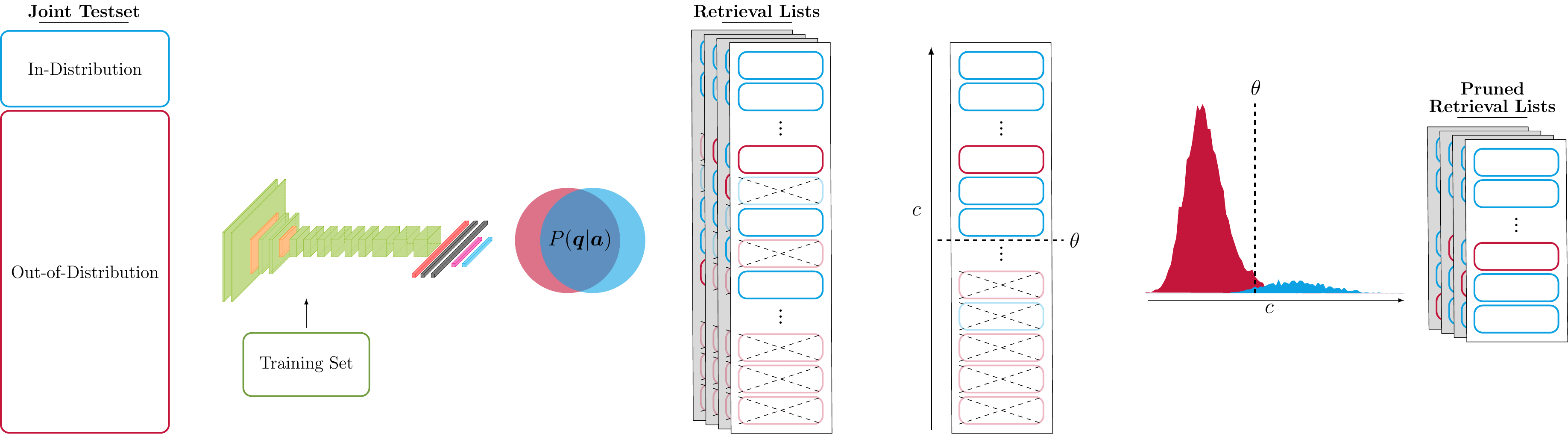}
	\caption{The figure visualizes the experimental setup.
		 A \textit{segmentation-based} word spotting system is trained on a training set and evaluated on a composed test set including \ID and \OD samples.
		 For each query, a retrieval list is obtained by evaluating a probabilistic retrieval model.
		 All samples are associated to a confidence measure.
		 Inaccurate estimations are removed by thresholding.}
	\label{fig:overview}
\end{figure*}

Word spotting is a powerful tool for exploring handwritten document collections.
Machine learning based methods got increasingly popular and showed exceptional performances on numerous academic benchmarks \cite{Giotis2017}.
It has been shown that attribute based word spotting systems are extremely robust to high variations in style and appearance of the documents \cite{Sudholt2018}.
While more and more sophisticated models emerge, they all share the assumption that representative training material is available.
This is often the case for an academic benchmark but in a real world application the system is faced with an unknown dataset.
Especially in historic and handwritten document collections, writing styles can change frequently.
Degradation can drastically change the visual appearance of the documents over the collection making them highly heterogeneous.
A limited amount of training material might be provided by some expert who annotated a small part of the dataset.
Assessing if the training material is representative is practically impossible without manually evaluating the entire document collection.

The inability to reject certain candidates is one of the main drawbacks of a retrieval system.
The result \wrt a given query is always an ordered list of instances.
A common retrieval system does not provide any information on which parts of a dataset it is able to retrieve reliable results.
State-of-the-art approaches make use of CNNs that learn the mapping between a word image and an attribute embedding \cite{Sudholt2018}.
Attributes are semantic entities shared between multiple classes, which have been shown to be highly robust representations and allow for zero-shot learning \cite{Almazan2014}.
Due to the overconfidence of neural networks, high attribute activations can be observed for whatever input is given to the network.
Even an image of random noise would result in an attribute representation and the retrieval system would rank it \wrt the query.
In order to extent the capabilities of a retrieval system, this work aims at finding a suitable confidence measure that allows to assess whether a CNN is able to estimate an accurate attribute representation.

We explore different confidence measures that will be evaluated according to the experimental setup depicted in \autoref{fig:overview}.
A \textit{segmentation-based} word spotting system is trained on an annotated set of training material.
In order to model a highly heterogenous dataset, the test set is composed of an in-distribution (\ID) part that is well represented by the training data and an out-of-distribution (\OD) part.
While not having any explicit relation to the training material, the OD set shall consist of a wide range of different samples, being more and less similar to the style of the training set.
The word spotting system is used to generate a retrieval list for each query over the composed test set.
Given a confidence measure, the generated retrieval lists are pruned by a simple thresholding method.
All samples above a given confidence should have an accurate attribute estimation without necessarily being from the \ID part.
Samples with an associated confidence below the threshold are rejected.
Thereby, inaccurate attribute estimations, which would lead to poor retrieval perfomance, are removed from the test set.
\section{Related Work}
\subsection{Word Spotting}
Word spotting describes the task of retrieving a subset of word images from a document collection that are relevant \wrt a query.
In contrast to methods aiming at directly transcribing a document, word spotting systems have been shown to be extremely robust.
This makes word spotting a highly suitable technique, especially when confronted with handwritten historic documents that often show high variability and suffer from degradation effects.
For an extensive overview of word spotting methods, see \cite{Giotis2017}.

In general, most approaches make several assumptions \wrt the document collection and the query protocol.
\textit{Segmentation-based} methods (\cf \eg \cite{Sudholt2018,Almazan2014}) require a previous segmentation of document pages into individual word images, which is in general not an easy to solve problem.
The \textit{segmentation-free} approach does not pose this requirement, but aims at solving the retrieval and segmentation problem jointly. 
Considering the query provided by the user, two different protocols are distinguished.
In case of \textit{query-by-example} (QbE), the query is provided as a word image.
\textit{Query-by-string} (QbS) allows for string based representations as queries. 

In \cite{Almazan2014}, the concept of attribute-based learning was introduced to the field of word spotting. 
Attributes are entities that are shared between different classes. 
With respect to word images, a specific word can be considered as a class while its characters can be interpreted as attributes.
Taking spatial relations between characters into account, the \textit{Pyramidal Histogram of Characters} (PHOC) is derived.
In \cite{Almazan2014}, the mapping between attribute embedding and word images is learned with a set of support vector machines.
This allows to map word images and strings in a common subspace where the retrieval problem can be solved by comparing distances between attribute vectors.
Inspired by the success of CNNs, \cite{Sudholt2018} used a neural network to learn the attribute embedding.
This approach outperformed all previous methods by a large margin and still defines the state-of-the-art for \textit{segmentation-based} word spotting.
In \cite{Rusakov2018}, a probabilistic retrieval model (PRM) is proposed.
While cosine similarity and euclidean distances do not provide a robust distance metric in high dimensional spaces, the PRM gives a probabilistic description of similarity between query and estimated attribute vector.
Even though the attribute CNN approach has shown excellent performance on numerous commonly used academic benchmarks, this comes at the cost of requiring training material.
Works such as \cite{Krishnan2016} and \cite{Gurjar2018} try to alleviate the data problem by transfer learning and incorporating synthetic data, but still the necessity of representative training data is inherent to any machine learning based approach.

\subsection{Uncertainty}
The task of uncertainty estimation of neural networks and \OD detection has recently been an active field.
Estimating the uncertainty of neural networks by applying dropout during test time was analysed by \cite{Gal2016}.
However, this method is computationally expensive, as one has to make a large number of forward passes through the network.
Different types of uncertainty were analysed by \cite{Harang2018}.
They distinguish model capacity uncertainty, intrinsic data uncertainty and open set uncertainty.

Approaches by \cite{Hendrycks2017} and \cite{DeVries2018} rely on additional surrogate models.
A baseline for the use of confidence measures for \OD detection in a classification scenario was proposed by \cite{Hendrycks2017}.
The use of the maximum softmax entry as a confidence measure yielded good results.
The interpretability however is a major drawback, as most neural networks are overconfident in their decisions.
This leads to high confidence values for most of the \OD samples.
They also suggested the use of a multi headed neural network utilizing an auxiliary decoder together with an 'abnormality module', which increased the separability between \ID and \OD examples significantly.
Multi headed neural networks were also used by \cite{DeVries2018} who added a second branch in parallel with the fully connected layer of a neural network.
During training, they used a joint loss function based on an interpolation approach between network prediction and supplied label.
The interpolation factor could then be interpreted as a confidence measure.
This yielded good performances on classification tasks but showed a high regularization effect, which made it necessary to introduce additional hyperparameters.
\section{Method}
With all proposed confidence measures, we aim at quantifying the quality of a predicted attribute vector.
The assumption is that data points that are dissimilar from the training distribution have an inaccurate attribute prediction, which results in a wrong position in the retrieval list. 

\subsection{Word Spotting}
\label{subsec:ws}

\begin{figure}
	\centering
	\input{gfx/phoconfnetID}
	\vspace{-1.5em}
	\caption{Task independent metaclassifier.}
	\label{fig:metaID}
\end{figure}
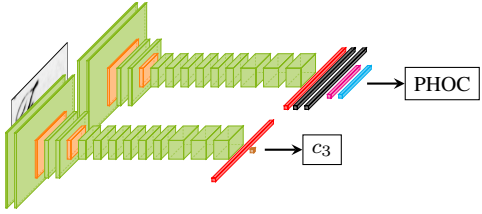
Our baseline word spotting system is based on the design of \cite{Sudholt2018}.
The attribute embedding is a 4-level PHOC representation of partitions $1$, $2$, $4$ and $8$ based on the lower case Latin alphabet plus digits. 
We employ a  TPP-PHOCNet to estimate a PHOC vector $\bm{\hat{a}}\in\left(0,1\right)^{540}$ for a word image.

In analogy to \cite{Rusakov2018} the retrieval list for the query $q$ with the corresponding attribute embedding $^{q}\bm{a}$ is ranked according to the posterior probability
\begin{equation*}
	p(^{q}\bm{a}|\bm{x}) = \prod_{i=1}^{540}\hat{a}_{i}^{^{q}a_i}\cdot(1-\hat{a}_i)^{(1-^{q}a_i)}
\end{equation*}
which serves as a similarity measure.
For a given annotation $t$, the quality of an estimated PHOC vector can be quantified by evaluating the probabilistic model \wrt the ground truth attribute embedding $^{t}\bm{a}$.
Therefore, $p_{t|x} = p(^{t}\bm{a}|\bm{x})$ describes the probability of $^t\bm{a}$ being the embedding of $\bm{x}$.

\subsection{Sigmoid Activation and Test Dropout}
\label{subsec:sigtdo}
Considering sigmoid activation as a pseudo probability, we derive the confidence measure $c_1$ by taking the mean over the activation of all active attributes. 
An attribute is considered active in case of $\hat{a}_i > 0.5$.

The second confidence measure is based on dropout at test time.
We apply dropout with a probability of $0.5$ to all but the last fully connected layer.
Each sample is passed through the network 100 times with both dropout layers being active.
The variance of the estimations for each attribute is determined.
The confidence measure $c_2$ is then obtained by averaging over all attribute variances.
Opposed to the other confidence measures, a high confidence corresponds to a small value of $c_2$.

\subsection{Task Independent Metaclassifier}
\label{subsec:metaID}
A task independent (TI) metaclassifier is an additional surrogate model, which has no relation of the task learned by the main model.
It receives the same input and classifies in \ID and \OD. For training, one only needs the training set of the main model and some data of a different distribution.
This could be anything from gaussian-/normal-noise over synthetically generated to real world samples.
This makes it rather easy to obtain the \OD data.
The proposed independent metaclassifier is shown in \autoref{fig:metaID}.
Its architecture is a replication of the PHOCNet, but with a different MLP part.
The original fully connected layers and the sigmoid output are replaced by a single neuron with sigmoid activation.

Let $X$ be the training set of word images for our task with a set of known PHOC vector representations $A=\{\bm{a}\in\{0,1\}^{540}\}$ and $O$ a set of word images, which are sampled from a different distribution than $X$, without a known transcription or PHOC vector.
The PHOCNet is trained on $X$ with labels $A$ to approximate the distribution $p(\bm{a}|\bm{x})$ whereas the metaclassifier is trained on $X\cup O$ to approximate the distribution $p(d|\bm{x})$, giving the probability that $\bm{x}$ belongs to $X$.
The label set $D$ for the training of the metaclassifier can be obtained as 

\begin{equation*}
	D=\{d(\bm{x}) | \bm{x}\in X\cup O\} \text{ with } d(\bm{x})=\begin{cases}
	1 & \text{if } \bm{x}\in X\\
	0 & \text{else}
	\end{cases}.
\end{equation*}

The surrogate model is trained with binary crossentropy loss and hyperparameters as described in \autoref{subsec:training}.
The confidence measure $c_3$ results from the penultimate layer of the metaclassifier.

\subsection{Task Dependent Metaclassifier}
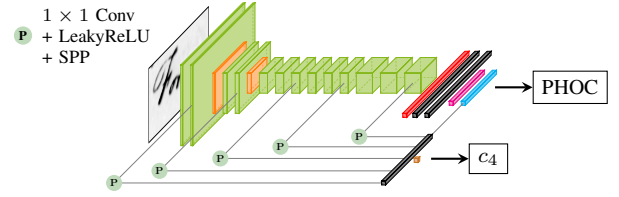
\begin{figure}
	\centering
	\input{gfx/phoconfnet}
	\vspace{-1.5em}
	\caption{Task dependent metaclassifier using deep features.}
	\label{fig:metaCl}
\end{figure}

\begin{figure*}
	\centering
	\input{plots/gw_bot_hist_overview}
	\vspace{-1.8em}
	\caption{Distribution of confidences of the ID (blue) and OD (red) set for all proposed methods.}
	\label{fig:gw_bot_histograms}
\end{figure*}
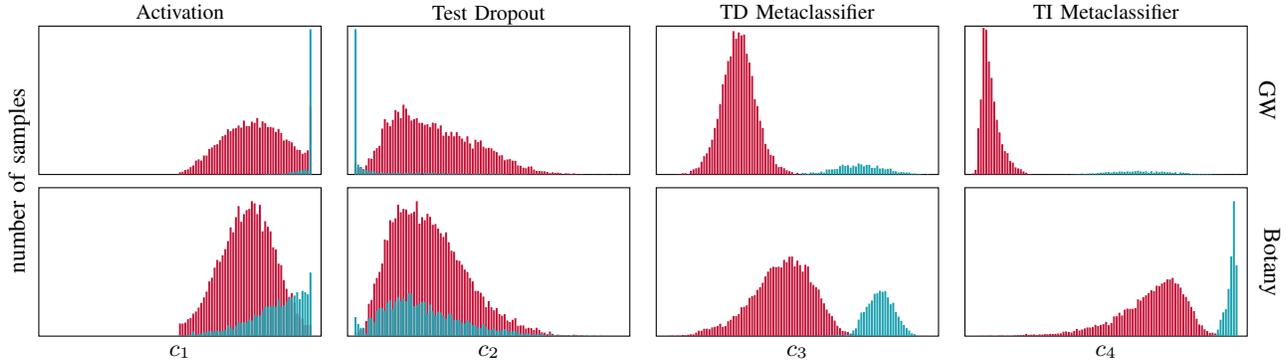
A task dependent (TD) metaclassifier receives the featuremaps learned by the PHOCNet as input.
Using the already learned representations, it allows to reduce the number of additional parameters and produces a confidence measure, which is semantically related to the main task.
Defining with $f_i := f_i(\bm{x},\bm{w})$ the output of the PHOCNet up until the $i$-th layer \wrt the weights $\bm{w}$, the task dependent metaclassifier learns to approximate the distribution $p(d|f_{s_1},\ldots, f_{s_l}),\quad s_1,\ldots,s_l\in\{1,\ldots,L\}$ with $L$ being the number of layers (only counting layers that have trainable weights).
Here, we choose $f_2$, $f_4$, $f_7$, $f_{10}$, $f_{13}$ and $f_{16}$, which are passed through a $1\times 1$ convolution with one feature map followed by a leaky ReLU activation and spatial pyramid pooling (SPP) up to level 4.
The resulting feature vectors and the penultimate layer of the PHOCNet are then concatenated and fed through a single neuron with sigmoid activation.
First, the PHOCNet weights are trained as described in \autoref{subsec:training}.
Afterwards, we freeze the weights of the attribute CNN and train the metaclassifier with binary crossentropy loss on $X\cup O$, label set $D$ and hyperparameters as described in \autoref{subsec:training}.
The confidence measure $c_4$ is then taken from the penultimate layer of the metaclassifier.
\section{Experiments}
See \autoref{fig:overview} for an overview of our experimental setup.
We train a \textit{segmentation-based} word spotting system on a designated training set.
The different confidence measures are then evaluated on a composed test set with the aim to distinguish between \ID and \OD and to prune inaccurate attribute vector estimations from the resulting retrieval lists.
\subsection{Datasets}
\begin{table}
	\centering
	\input{tables/datasets}
	\caption{Datasets used in this work. 
		 Each dataset is either used to train the PHOCNet, the metaclassifier (MC) or models ID or OD.}
	\label{tab:datasets}
\end{table}
We use four different publicly available datasets.
The George Washington (GW) and Botany dataset are well known to the word spotting community and both have a distinctive style.
Our baseline word spotting system is trained on the training partition in case of Botany or follows the common four-fold cross validation approach of GW, \cf \cite{Almazan2014}.
The respective test partitions are used as ID sets.
Note that for the Botany test set, annotations are only available for those samples which are relevant \wrt a query following the standard protocol \cite{Pratikakis2016}.
The IAM database contains a wide range of different writing styles, which makes it a suitable choice as the \OD set.
Our choice of datasets is further motivated by the fact that our experiments require overlapping lexica, since word images relevant to a query shall exist in the \ID and \OD set.
For training the metaclassifiers, we use the synthetic dataset HWSynth to model \OD samples.
See \autoref{tab:datasets} for an overview of the different datasets.


\subsection{Training Setup}
\label{subsec:training}
For all our experiments, we train the TPP-PHOCNet for $100\,000$ iterations with an initial learning rate of $10^{-4}$, which is divided by $10$ after $70\,000$ iterations.
We use Adam optimization with a mini-batch size of $10$, hyperparameters $\beta_1=0.9$ and $\beta_2=0.999$ and a weight decay of $5\cdot10^{-5}$.
Analogue to \autoref{subsec:sigtdo}, we apply two dropout layers during training.
Furthermore, a simple data augmentation strategy is used as we apply a random affine transformation to all input images at training time.

The metaclassifiers are trained for $25\,000$ iterations with an initial learning rate of $10^{-2}$, which is divided by $10$ after $10\,000$, $15\,000$ and $20\,000$ iterations.
For optimization, we use the same Adam optimizer as for the TPP-PHOCNet with a weight decay of $5\cdot 10^{-4}$.

\subsection{Results and Discussion}
Our experiments investigate the following questions:
\begin{itemize}
	\item Are the proposed confidence measures suitable for \textit{separating} \ID and \OD examples?
	\item Does a relation between confidence and the \textit{quality} of an estimated attribute vector exist?
	\item Is it possible to prune the resulting retrieval lists by \textit{thresholding} in order to compromise between the reliability of the results and the coverage of the dataset?
\end{itemize}

\textbf{\textit{Separability}}: According to our experimental setup, we assume that the word spotting system is giving reliable results on the \ID sets.
This does not hold true for the \OD part, as no related training material was used.
The proposed confidence measures are supposed to quantify this assumption and they should yield higher confidences for the \ID parts.
\autoref{fig:gw_bot_histograms} shows the distributions of the different confidence measures.
On GW, high confidences can be observed \wrt to the attribute estimations of the \ID samples and the confidence measures derived from sigmoid activation or via test dropout.
In case of Botany, this observation does not hold true.
For sigmoid activation, slightly higher confidences can be observed for the \ID set.
The distributions of confidences resulting from test dropout do not allow to easily distinguish between \ID and \OD.
In contrast, both metaclassifiers are almost perfectly able to distinguish between \ID and \OD samples on both datasets.
The confidences for \ID samples are almost exclusively higher than those of the \OD samples.

\textbf{\textit{Relation to Attribute Vector Quality}}: As discussed in \autoref{subsec:ws}, the quality of an attribute vector estimation \wrt the ground truth transcription can be described by $p_{t|x}$.
Even though the metaclassifiers are almost perfectly able to distinguish \ID and \OD, that is not exactly what we aim for in terms of measuring confidence.
Due to the multi writer characteristic of the \OD set, a wide range of writing styles is represented.
Given the training material, a confidence measure should quantify how well an attribute vector can be estimated for a sample.
In case of the \OD set, this should correspond to the quality of an estimated attribute vector measured by the posterior $p_{t|x}$.
\begin{figure}
	\centering
	\input{plots/confnet_scatter}
	\vspace{-2.4em}
	\caption{Distribution of ID and OD samples over the negative logarithmic posterior $p_{t|x}$ and the confidence measure.
		 }
	\label{fig:gw_bot_scatter}
\end{figure}
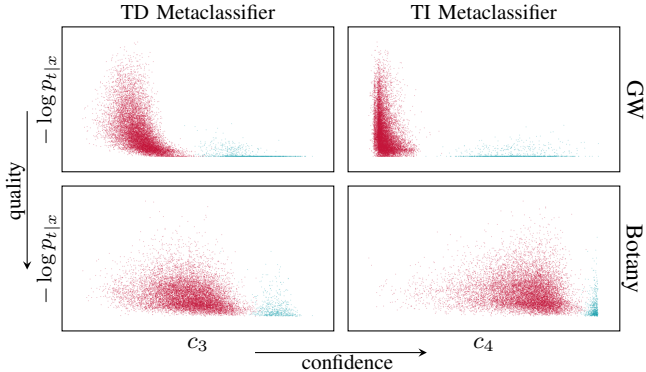

\autoref{fig:gw_bot_scatter} shows the distribution of \ID and \OD samples over the confidence measure and the posterior $p_{t|x}$, which quantifies the quality of the estimation.
Almost all \ID samples of GW have a high probability $p_{t|x}$ and can be assumed to be precise estimations of their ground truth embeddings.
This is not the case for the Botany dataset, where samples with a high confidence are not concentrated at high posteriors.
Despite a high mean average precision (mAP) on Botany according to the standard protocol, this observation indicates that the task of estimating an attribute vector is solved poorly compared to the GW dataset.
Comparing the task dependent and independent metaclassifiers, one can observe a correlation between confidence and quality especially for task-dependency.
This is far less observable for the task independent metaclassifier.
While the deep features used by the task dependent metaclassifier provide a semantic relation to the task of estimating an attribute vector, the task independent confidence measure is solely based on visual similarity.
This might explain the observed characteristics.
\begin{figure}
	\centering
	\input{plots/wercurves}
	\caption{Cumulative word error rates.
		 }
	\label{fig:wercurves}
\end{figure}
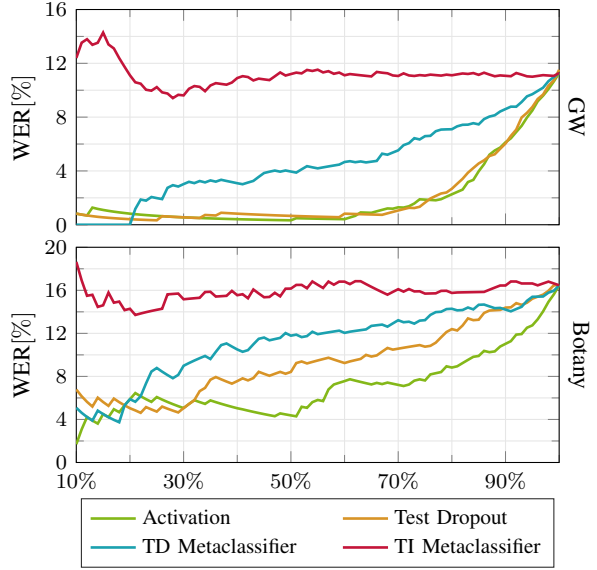

To further investigate the relation between confidence measures and attribute vector quality, we conduct the following experiment.
The given word spotting system can easily be extended to perform lexicon-based word recognition analogue to the approach of \cite{Poznanski2016}.
Let $\mathbb{L}$ be the lexicon obtained from all available training and test set transcriptions with corresponding attribute representation $^{l}\bm{a}$.
Based on the attribute representation estimated for a word image $\bm{x}$ the recognition result $s$ is given by
\begin{equation*}
	s = \argmax_{l \in \mathbb{L}} p\left(^{l}\bm{a}|\bm{x}\right).
\end{equation*}
Note that we do not expect the proposed system to give state-of-the-art recognition results, as we believe that for this task sequential models are superior to holistic attribute based approaches.
Nevertheless, evaluating word error rates gives a performance measure that describes if the most probable string of the lexicon is equal to the ground truth transcription.
This experiment allows to evaluate the probability $p_{t|x}$ for a given attribute vector estimation, with a commonly known and easier to interpret performance measure.

We first sort the \ID set \wrt confidence.
Then we determine the word error rate (WER) for the most confident $x\%$ of the dataset.
\autoref{fig:wercurves} shows the WER over different portions of the \ID datasets.
Even though our system yields a WER over the entire GW test set of $11.52\%$, it drops to below $2\%$ for the most confident $70\%$ using sigmoid activations or test dropout as confidence measure.
All confidence measures, despite the task independent metaclassifier, achieve significantly lower WER on the more confident parts of the test sets.
This further supports our conclusion that those confidence measures, semantically related to the task, also quantify the quality of an estimated attribute vector.

\begin{figure}
	\centering
	\input{plots/gw0005}
	\vspace{-2em}
	\caption{Thresholding characteristics of a task dependent metaclassifier on a test set composed of GW and the IAM database.}
	\label{fig:thresholding}
\end{figure}
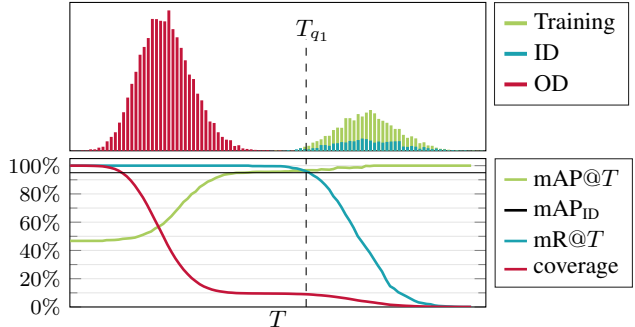

\textbf{\textit{Thresholding}}: \autoref{fig:thresholding} visualizes the distribution of confidences for the training, \ID and \OD set, given a task dependent metaclassifier \wrt the GW dataset.
If only \ID samples are considered, the word spotting system produces the baseline performance $\text{mAP}_\text{ID}$ for QbS word spotting.
All unique strings of the \ID set serve as queries ones. 
For the composed test set, we consider the case where all samples below a certain confidence are removed from the retrieval lists.
At each possible threshold $T$, the $\mapT$ gives the QbS mAP over the pruned dataset.
Additionally, \autoref{fig:thresholding} reports the $\mrT$, which is the mean recall over the pruned dataset.
The coverage describes the percentage of the joint test set lying above the threshold $T$.

Starting with the entire joint test set, the $\mapT$ is significantly below the $\text{mAP}_\text{ID}$, since poor attribute estimations are not ranked properly.
Increasing the threshold leads to removing inaccurate estimations, improving the $\mapT$ at the cost of a lower coverage.
The $\mapT$ approaches the baseline performance as most of the \OD samples are removed from the retrieval lists.
A further increase of the threshold leads to pruning \ID samples, which lowers the $\mrT$.
As we only consider queries that occur at least once, the $\mapT$ further improves as only increasingly confident parts of the \ID set are considered.
The threshold can be considered a parameter allowing to compromise between the quality of retrieval results and the coverage of the test set.
It is also possible to estimate a threshold based on the training distribution.
As depicted in \autoref{fig:thresholding} top, using the one percent quantile of the training distribution $T_{q_{1}}$ as a threshold, the \ID set is almost exactly separated from the \OD set, reproducing the baseline performance on the pruned test set.

\section{Conclusions}
In this work, we proposed four different confidence measures for a word spotting system and showed that those semantically related to the task provide an estimation of the attribute vector quality.
We conclude that a task dependent metaclassifier is a suitable model to distinguish ID and OD samples while quantifying quality.
This allows to identify parts of a dataset for which reliable results can be obtained.

\bibliographystyle{IEEEtran}
\bibliography{main}
\end{document}

%% file: macros.tex
\DeclareMathOperator*{\argmax}{argmax}

\newcommand{\mapT}{\text{mAP}@T}
\newcommand{\mrT}{\text{mR}@T}

\newcommand{\eg}{e.g.\xspace}
\newcommand{\cf}{c.f.\xspace}
\newcommand{\wrt}{w.r.t.\xspace}
\newcommand{\ID}{ID\xspace}
\newcommand{\OD}{OD\xspace}


%% file: gfx/phoconfnetID.tex
\begin{tikzpicture}
	\node[anchor=north west,rectangle, minimum width=0.48\textwidth, minimum height=0.13\textheight] (main) at (0,0) {};

	\node[left=1em, anchor=north,rectangle] (phoconfnet) at (main.north) 
	{\includegraphics[height=0.12\textheight]{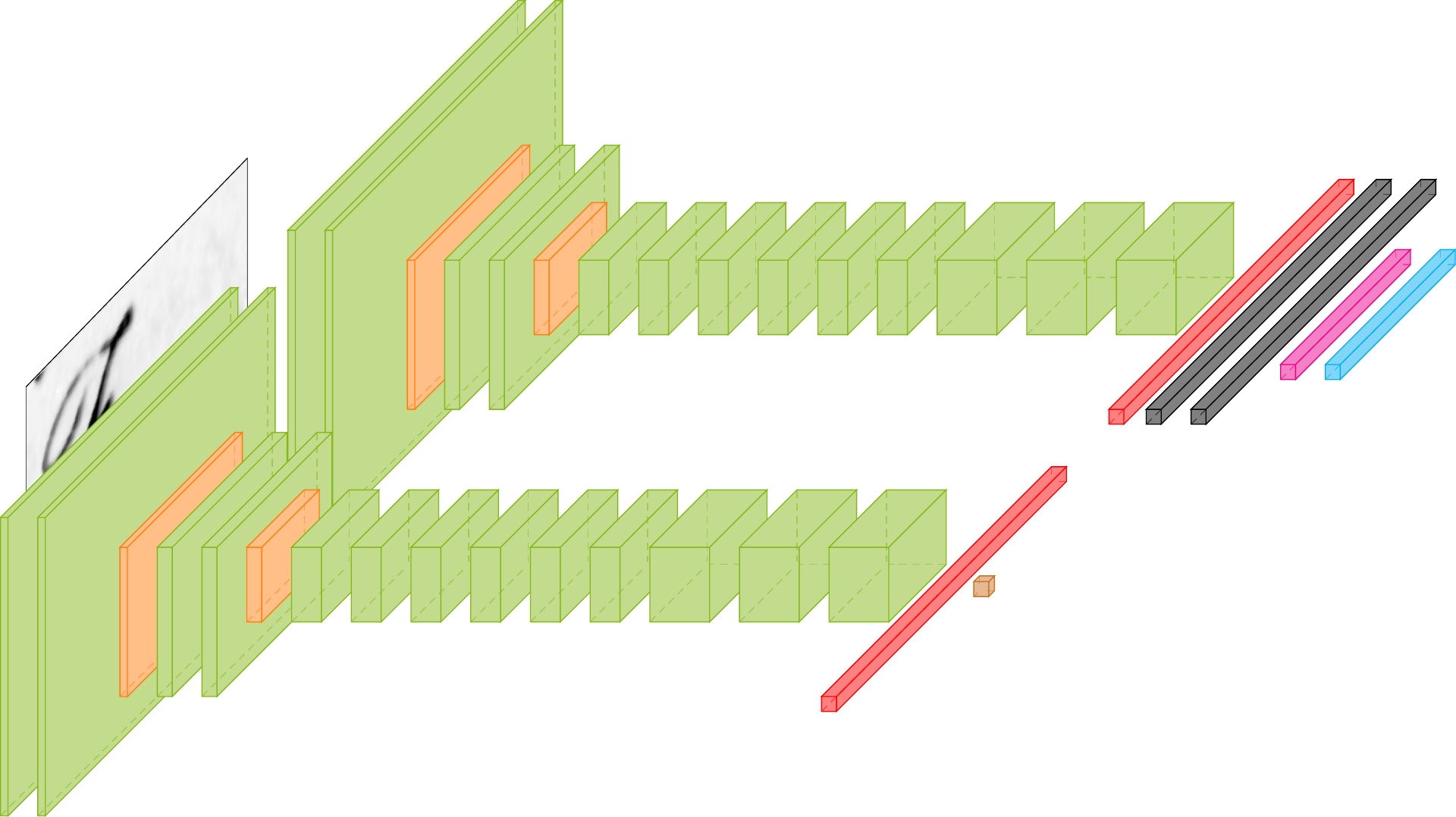}};

	\path[thick, draw,->] (6.3, -1.2) -- (6.8, -1.2) node[rectangle, draw, anchor=west] (phoc) {\footnotesize PHOC};
	\path[thick, draw,->] (4.94, -2.08) -- (5.44, -2.08) node[rectangle, draw, anchor=west] (c) {\footnotesize $c_3$};
\end{tikzpicture}

%% file: gfx/phoconfnet.tex
\begin{tikzpicture}
	\node[anchor=north west,rectangle, minimum width=0.48\textwidth, minimum height=0.13\textheight] (main) at (0,0) {};

	\node[left=0em, anchor=north,rectangle] (phoconfnet) at (main.north) 
	{\includegraphics[height=0.11\textheight]{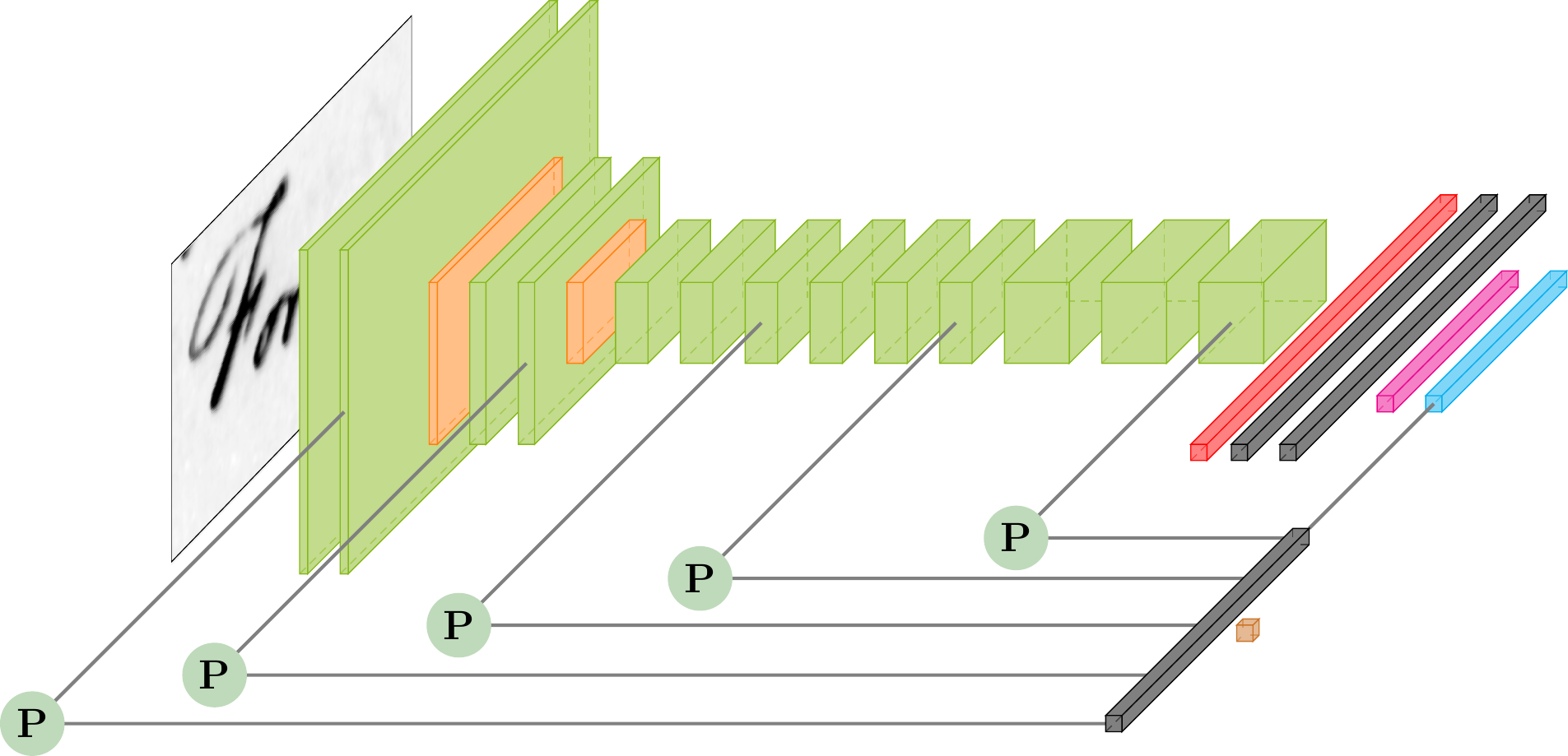}};

	\path[thick, draw,->] (6.85, -1.25) -- (7.35, -1.25) node[rectangle, draw, anchor=west] (phoc) {\footnotesize PHOC};
	\path[thick, draw,->] (6, -2.2) -- (6.5, -2.2) node[rectangle, draw, anchor=west] (c) {\footnotesize $c_4$};


	\node[below right=0.2em and 2em, anchor=north west] (legend) at (main.north west) {\parbox{4.2cm}{\scriptsize $1\times 1$ Conv\\ \scriptsize+ LeakyReLU\\ \scriptsize+ SPP}};
	\node[left=0em, anchor=east, circle, fill=tucol-kuehl4, minimum size=0.1mm, inner sep=0.3mm] at (legend.west) {\tiny\bfseries P};

\end{tikzpicture}

%% file: plots/gw_bot_hist_overview.tex
\begin{tikzpicture}
	
	\node[rectangle, minimum width=\textwidth, minimum height=0.27\textwidth] at (0,0) (main) {};

	\node[right=1em of main.west,anchor=north, rotate=90] {number of samples};

	\coordinate[below right=1em and 2.5em of main.north west, anchor=north west](gw1){};
	\begin{axis}[
		name=gw_hist1,
		at=(gw1), anchor=north west,
		width=0.3\textwidth,
		height=0.2\textwidth,
		xmin=-3,
		xmax=103,
		ytick=\empty,
		xtick=\empty,
		ymin=0,
		ymax=1000,
		ybar,   
		bar width=0.75pt,
		bar shift=0,
		]

		\addplot[tucol-rot2,fill,draw opacity=0] table[x index=0, y index=2, col sep=comma] 
		{data/plot_gw1001_iam_activation_hist.csv};	

		\addplot[tucol-kraeftig3,fill,draw opacity=0] table[x index=0, y index=1, col sep=comma] 
		{data/plot_gw1001_iam_activation_hist.csv};
	\end{axis}

	\node[above=-0.1em of gw_hist1] {\footnotesize{Activation}};

	\coordinate[right=1em of gw_hist1.south east](gw2){};
	\begin{axis}[
		name=gw_hist2,
		at=(gw2), anchor=south west,
		width=0.3\textwidth,
		height=0.2\textwidth,
		xtick=\empty,
		xmin=-3,
		xmax=103,
		ytick=\empty,
		ymin=0,
		ymax=750,
		ybar,   
		bar width=0.75pt,
		bar shift=0,
		]

		\addplot[tucol-rot2,fill,draw opacity=0] table[x index=0, y index=2, col sep=comma] 
		{data/plot_gw1001_iam_a1005050_hist.csv};	
		\addplot[tucol-kraeftig3,fill,draw opacity=0] table[x index=0, y index=1, col sep=comma] 
		{data/plot_gw1001_iam_a1005050_hist.csv};

	\end{axis}

	\node[above=-0.3em of gw_hist2] {\footnotesize{Test Dropout}};

	\coordinate[right=1em of gw_hist2.south east](gw3){};
	\begin{axis}[
		name=gw_hist3,
		at=(gw3), anchor=south west,
		width=0.3\textwidth,
		height=0.2\textwidth,
		xtick=\empty,
		xmin=-3,
		xmax=103,
		ytick=\empty,
		ymin=0,
		ymax=800,
		ybar,   
		bar width=0.75pt,
		bar shift=0,
		]

		\addplot[tucol-rot2,fill,draw opacity=0] table[x index=0, y index=2, col sep=comma] 
		{data/plot_gw0007_iam_phoconfnet_hist.csv};	

		\addplot[tucol-kraeftig3,fill,draw opacity=0] table[x index=0, y index=1, col sep=comma] 
		{data/plot_gw0007_iam_phoconfnet_hist.csv};
	\end{axis}

	\node[above=-0.1em of gw_hist3] {\footnotesize{TD Metaclassifier}};

	\coordinate[right=1em of gw_hist3.south east](gw4){};
	\begin{axis}[
		name=gw_hist4,
		at=(gw4), anchor=south west,
		width=0.3\textwidth,
		height=0.2\textwidth,
		xtick=\empty,
		xmin=-3,
		xmax=103,
		ytick=\empty,
		ymin=0,
		ymax=1800,
		ybar,   
		bar width=0.75pt,
		bar shift=0,
		]
		
		\addplot[tucol-rot2,fill,draw opacity=0] table[x index=0, y index=2, col sep=comma] 
		{data/plot_gw0006_iam_phoconfnetID_hist.csv};	

		\addplot[tucol-kraeftig3,fill,draw opacity=0] table[x index=0, y index=1, col sep=comma] 
		{data/plot_gw0006_iam_phoconfnetID_hist.csv};
	\end{axis}

	\node[above=-0.1em of gw_hist4] {\footnotesize{TI Metaclassifier}};

	\node[right=0.1em of gw_hist4.east, anchor=south, rotate=270] {GW};

	\coordinate[below=0.5em of gw_hist1.south west](bot1){};
	\begin{axis}[
		name=bot_hist1,
		at=(bot1), anchor=north west,
		width=0.3\textwidth,
		height=0.2\textwidth,
		xtick=\empty,
		xlabel={$c_1$},
		x label style={yshift=1.5em},
		xmin=-3,
		xmax=103,
		ytick=\empty,
		ymin=0,
		ybar,   
		bar width=0.75pt,
		bar shift=0,
		]          

		\addplot[tucol-rot2,fill,draw opacity=0] table[x index=0, y index=2, col sep=comma] 
		{data/plot_bot0032_iam_activation_hist.csv};	
                          
		\addplot[tucol-kraeftig3,fill,draw opacity=0] table[x index=0, y index=1, col sep=comma] 
		{data/plot_bot0032_iam_activation_hist.csv};

	\end{axis}         
                          
	\coordinate[right=1em of bot_hist1.south east](bot2){};
	\begin{axis}[      
		name=bot_hist2,
		at=(bot2), anchor=south west,
		width=0.3\textwidth,
		height=0.2\textwidth,
		xtick=\empty,
		xlabel={$c_2$},
		x label style={yshift=1.5em},
		xmin=-3,
		xmax=103,
		ytick=\empty,
		ymin=0,   
		ybar,   
		bar width=0.75pt,
		bar shift=0,
		]         

		\addplot[tucol-rot2,fill,draw opacity=0] table[x index=0, y index=2, col sep=comma] 
		{data/plot_bot0032_iam_a1005050_hist.csv};	
                          
		\addplot[tucol-kraeftig3,fill,draw opacity=0] table[x index=0, y index=1, col sep=comma] 
		{data/plot_bot0032_iam_a1005050_hist.csv};

	\end{axis}        
                          
	\coordinate[right=1em of bot_hist2.south east](bot3){};
	\begin{axis}[     
		name=bot_hist3,
		at=(bot3), anchor=south west,
		width=0.3\textwidth,
		height=0.2\textwidth,
		xtick=\empty,
		xlabel={$c_3$},
		x label style={yshift=1.5em},
		xmin=-3,
		xmax=103,
		ytick=\empty,
		ymin=0,   
		ymax=800, 
		ybar,   
		bar width=0.75pt,
		bar shift=0,
		]         
                         
		\addplot[tucol-rot2,fill,draw opacity=0] table[x index=0, y index=2, col sep=comma] 
		{data/plot_bot0002_iam_phoconfnet_hist.csv};	
		
		\addplot[tucol-kraeftig3,fill,draw opacity=0] table[x index=0, y index=1, col sep=comma] 
		{data/plot_bot0002_iam_phoconfnet_hist.csv};
	\end{axis}        
                          
	\coordinate[right=1em of bot_hist3.south east](bot4){};
	\begin{axis}[     
		name=bot_hist4,
		at=(bot4), anchor=south west,
		width=0.3\textwidth,
		height=0.2\textwidth,
		xtick=\empty,
		xlabel={$c_4$},
		x label style={yshift=1.5em},
		xmin=-3,
		xmax=103,  
		ytick=\empty,
		ymin=0,   
		ybar,   
		bar width=0.75pt,
		bar shift=0,
		]         
                         
		\addplot[tucol-rot2,fill,draw opacity=0] table[x index=0, y index=2, col sep=comma] 
		{data/plot_bot0009_iam_phoconfnetID_hist.csv};	

		\addplot[tucol-kraeftig3,fill,draw opacity=0] table[x index=0, y index=1, col sep=comma] 
		{data/plot_bot0009_iam_phoconfnetID_hist.csv};
	\end{axis}        

	\node[right=0.1em of bot_hist4.east, anchor=south, rotate=270] {Botany};
\end{tikzpicture}    

%% file: tables/datasets.tex
\begin{tabular}{c|c|c|c|c}
	Dataset		             & Train   & Test & Historic & Writers	\\ \hline	
	GW \cite{Rath2007}	     & PHOCNet & ID	 & yes 	    & 1		\\
	Botany \cite{Pratikakis2016} & PHOCNet & ID	 & yes 	    & 1		\\
	IAM \cite{Marti2002} 	     & -	  & OD	 & no 	    & 657	\\
	HWSynth \cite{Krishnan2016}  & MC 	  & -	 & no	    & 0		\\
\end{tabular}

%% file: plots/confnet_scatter.tex
\begin{tikzpicture}
	\node[anchor=north west,rectangle, minimum width=0.48\textwidth, minimum height=0.3\textwidth] (main) at (0,0) {};
	
	\coordinate[below right=1.4em and 2.3em of main.north west, anchor=north west](plot_anchor){};

	\node[anchor=north west,rectangle,draw] (gw_td) at (plot_anchor) 
	{\includegraphics[width=0.19\textwidth]{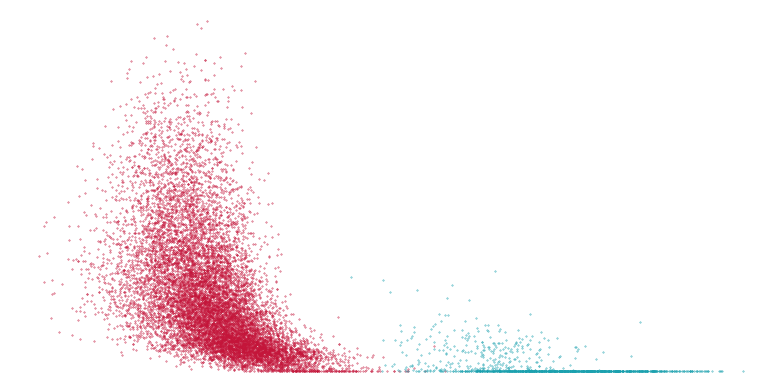}};

	\node[right=0.5em of gw_td.east,rectangle,draw] (gw_ti)
	{\includegraphics[width=0.19\textwidth]{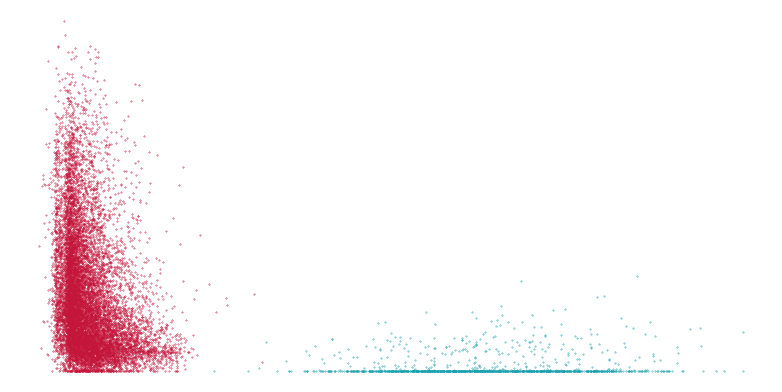}};

	\node[below=0.5em of gw_td.south west,anchor=north west,rectangle,draw] (bot_td) 
	{\includegraphics[width=0.19\textwidth]{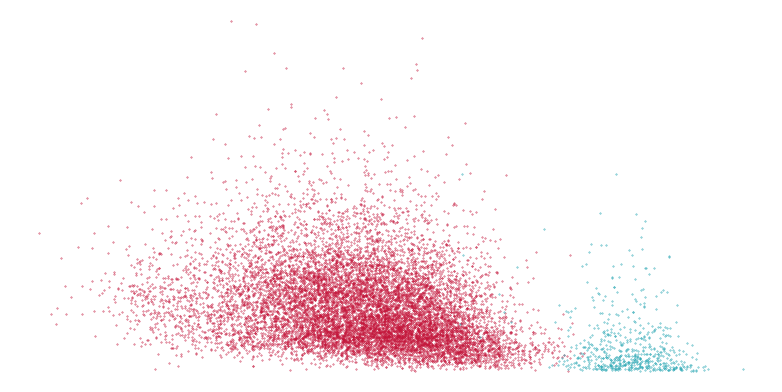}};

	\node[right=0.5em of bot_td.east,rectangle,draw] (bot_ti)
	{\includegraphics[width=0.19\textwidth]{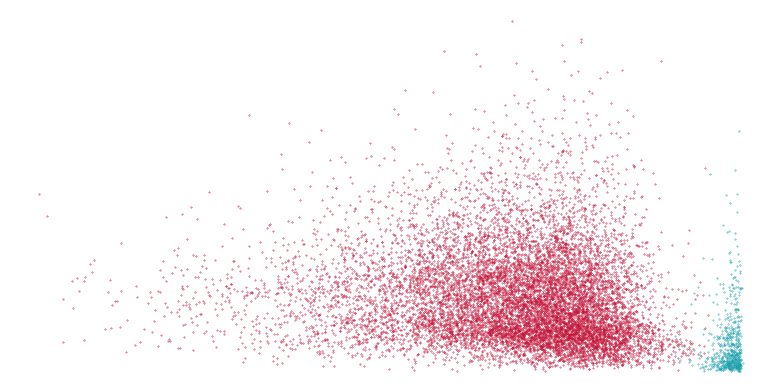}};
	
	\node[right=-0.15em of main.west, anchor=north, rotate=90] {\footnotesize{quality}};
	\coordinate[below right=3em and 1em of main.west] (qarrow_end) {};	
	\coordinate[above right=3em and 1em of main.west] (qarrow_start) {};
	\draw[->] (qarrow_start) -> (qarrow_end);

	\coordinate[below left=0.8em and 3em of bot_td.south east] (carrow_start) {};
	\coordinate[below right=0.8em and 3em of bot_ti.south west] (carrow_end) {};
	\draw[->] (carrow_start) -> (carrow_end);
	\node[below right=1.8em and 0.5em of bot_td.south east, anchor=south] {\footnotesize{confidence}};

	\node[left=-0.3em of gw_td.west, anchor=south, rotate=90] {\footnotesize{$-\log p_{t|x}$}};
	\node[left=-0.3em of bot_td.west, anchor=south, rotate=90] {\footnotesize{$-\log p_{t|x}$}};

	\node[above=-0.1em of gw_td] {\footnotesize{TD Metaclassifier}};
	\node[above=-0.1em of gw_ti] {\footnotesize{TI Metaclassifier}};
	\node[right=-0.1em of gw_ti.east, anchor=south,rotate=270] {GW};
	\node[right=-0.3em of bot_ti.east, anchor=south,rotate=270] {Botany};

	\node[below=-0.1em of bot_td.south, anchor=north] {$c_3$};
	\node[below=-0.1em of bot_ti.south, anchor=north] {$c_4$};
\end{tikzpicture}

%% file: plots/wercurves.tex
\begin{tikzpicture}
	\node[anchor=north west,rectangle, minimum width=0.48\textwidth, minimum height=0.435\textwidth] (main) at (0,0) {};
	\coordinate[below right=0.4em and 1em of main.north] (plot_anchor) {};
	\begin{axis}[
		name=plot_gw,
		at=(plot_anchor.north), anchor=north,
		width=0.45\textwidth,
		height=0.25\textwidth,
		xmin=10,
		xmax=100,
		xtick={10,30,50,70,90},
		xticklabels=\empty,
		ytick={0,0.04,0.08,0.12,0.16},
		ylabel={$\text{WER} [\%]$},
		y label style={yshift=-1.5em},
		yticklabels={\footnotesize{$0$},\footnotesize{$4$}, \footnotesize{$8$},
			     \footnotesize{$12$},\footnotesize{$16$}},
		minor tick num=1,
		ymin=0,
		ymax=0.16,
		grid=both,
		grid style={black!10}
		]

		\addplot[tucol-kraeftig1,mark=none,line width=1pt] table[x index=0, y index=1, col sep=comma] 
		{data/plot_gw1001_gw_activation_wer.csv};
		\addplot[tucol-kraeftig2,mark=none,line width=1pt] table[x index=0, y index=1, col sep=comma] 
		{data/plot_gw1001_gw_test_dropout_wer.csv};


		\addplot[tucol-kraeftig3,mark=none,line width=1pt] table[x index=0, y index=1, col sep=comma] 
		{data/plot_gw0005_gw_phoconfnet_wer.csv};
		\addplot[tucol-rot2,mark=none,line width=1pt] table[x index=0, y index=1, col sep=comma] 
		{data/plot_gw0004_gw_phoconfnetID_wer.csv};
	\end{axis}
	\node[right=0em of plot_gw.east, rotate=270, anchor=south] {GW};

	\node [below=0.5em of plot_gw.south west] (bot_wer) {};
	\begin{axis}[
		name=plot_bot,
		at=(bot_wer), anchor=north west,
		width=0.45\textwidth,
		height=0.25\textwidth,
		xmin=10,
		xmax=100,
		xtick={10,30,50,70,90},
		xticklabels={\footnotesize{$10\%$},\footnotesize{$30\%$},\footnotesize{$50\%$},
			     \footnotesize{$70\%$},\footnotesize{$90\%$}},
		ytick={0,0.04,0.08,0.12,0.16,0.2},
		yticklabels={\footnotesize{$0$},\footnotesize{$4$},\footnotesize{$8$},
			     \footnotesize{$12$},\footnotesize{$16$},\footnotesize{$20$}},
		ylabel={$\text{WER} [\%]$},
		y label style={yshift=-1.5em},
		yticklabel pos=left,
		minor tick num=1,
		ymin=0,
		ymax=0.20,
		grid=both,
		grid style={black!10}
		]	

		\addplot[tucol-kraeftig1,mark=none,line width=1pt] table[x index=0, y index=1, col sep=comma] 
		{data/plot_bot0032_bot_activation_wer.csv};
		\addplot[tucol-kraeftig2,mark=none,line width=1pt] table[x index=0, y index=1, col sep=comma] 
		{data/plot_bot0032_bot_test_dropout_wer.csv};
		

		\addplot[tucol-kraeftig3,mark=none,line width=1pt] table[x index=0, y index=1, col sep=comma] 
		{data/plot_bot0001_bot_phoconfnet_wer.csv};
		\addplot[tucol-rot2,mark=none,line width=1pt] table[x index=0, y index=1, col sep=comma] 
		{data/plot_bot0007_bot_phoconfnetID_wer.csv};
	\end{axis}
	\node[right=0em of plot_bot.east, rotate=270, anchor=south] {Botany};


    \node[below=1em of plot_bot.south west] (node_legend) {};
    \begin{axis}[%
    at={(node_legend)}, anchor=south west,
    hide axis,
    width=0.3\textwidth,
    height=0.1\textheight,
    xmin=0, xmax=1,
    ymin=0, ymax=1,
    legend columns=2,
    legend style={
        draw=white!15!black,
        legend cell align=left,
        /tikz/column 2/.style={
            column sep=15pt,
        },
        at={(0.855,0)},
        anchor=north
    },    
    ]
    	\addlegendimage{tucol-kraeftig1,mark=none, line width=1pt}
	\addlegendentry{\footnotesize{Activation}};

	\addlegendimage{tucol-kraeftig2,mark=none,line width=1pt}
	    \addlegendentry{\footnotesize{Test Dropout}};
    	
	\addlegendimage{tucol-kraeftig3,mark=none,line width=1pt}
	\addlegendentry{\footnotesize{TD Metaclassifier}};
	
	\addlegendimage{tucol-rot2,mark=none,line width=1pt}
	\addlegendentry{\footnotesize{TI Metaclassifier}};
    \end{axis}

\end{tikzpicture}

%% file: plots/gw0005.tex
\begin{tikzpicture}
	\node[anchor=north west,rectangle, minimum width=0.48\textwidth, minimum height=0.24\textwidth] (main) at (0,0) {};
	\coordinate[right=2.5em of main.north west, anchor=north west](plot_anchor){};
	\begin{axis}[
		name=gw_hist,
		at=(plot_anchor), anchor=north west,
		width=0.4\textwidth,
		height=0.2\textwidth,
		xtick=\empty,
		xlabel=\empty,
		x label style={yshift=1.5em},
		xmin=-5000,
		xmax=5000,
		ytick=\empty,
		y label style={yshift=-2.5em},
		ymin=0,
		ymax=800,
		ybar,   
		bar width=1.1pt,
		bar shift=0,
		]

		\addplot[tucol-kraeftig1!70,fill,draw opacity=0] table[x index=0, y index=3, col sep=comma] 
		{data/plot_gw0005_iam_phoconfnet_hist.csv};
		
		\addplot[tucol-kraeftig3,fill,draw opacity=0] table[x index=0, y index=1, col sep=comma] 
		{data/plot_gw0005_iam_phoconfnet_hist.csv};
		
		\addplot[tucol-rot2,fill,draw opacity=0] table[x index=0, y index=2, col sep=comma] 
		{data/plot_gw0005_iam_phoconfnet_hist.csv};
	
		\draw[black,dashed] (568,0) -| (568,600);
		\node at (585, 650) {$T_{q_{1}}$};
	\end{axis}

    	\node[above left=0.45cm and -0.8cm of gw_hist.east] (node_legend) {};
    	\begin{axis}[%
    	at={(node_legend.east)}, anchor=west,
    	hide axis,
    	width=0.1\textwidth,
    	height=0.1\textheight,
    	xmin=0, xmax=1,
    	ymin=0, ymax=1,
    	legend columns=1,
    	legend style={
    	    draw=white!15!black,
    	    legend cell align=left,
    	    /tikz/column 2/.style={
    	        column sep=15pt,
    	    },
    	    at={(1.05, 1.1)},
    	    anchor=north
    	},    
	/pgfplots/legend image code/.code={	
		\draw[mark repeat=2,mark phase=2,#1] 
        	    plot coordinates {
                	(0cm,0cm) 
                	(0.1cm,0cm)
                	(0.2cm,0cm)
                	(0.3cm,0cm)
            	};
	}
    	]
    		\addlegendimage{tucol-kraeftig1!70,mark=none,line width=1.5pt}
    		\addlegendentry{Training};

    	    	\addlegendimage{tucol-kraeftig3,mark=none,line width=1.5pt}
    		\addlegendentry{ID};
    		
    	    	\addlegendimage{tucol-rot2,mark=none,line width=1.5pt}
    		\addlegendentry{OD};
    	\end{axis}

	\node[below=-0.8em of gw_hist.south west] (node_tcurve) {};
	\begin{axis}[
		name=gw_tcurve,
		at={(node_tcurve.south)}, anchor=above north west,
		width=0.4\textwidth,
		height=0.2\textwidth,
		xtick=\empty,
		xlabel={$T$},
		x label style={yshift=1.7em},
		xmin=-5000,
		xmax=5000,
		ytick={0,0.2,0.4,0.6,0.8,1},
		yticklabels={\footnotesize{$0\%$},\footnotesize{$20\%$},\footnotesize{$40\%$},
			     \footnotesize{$60\%$},\footnotesize{$80\%$},\footnotesize{$100\%$}},
		minor tick num=1,
		y label style={yshift=-2.5em},
		ymin=0,
		ymax=1.05,
		grid=both,
		grid style={black!10}]
		
		\addplot[tucol-kraeftig1!70,mark=none,line width=1] table[x index=0, y index=1, col sep=comma] 
		{data/plot_gw0005_phoconfnet_tcurve.csv};
		
		\addplot[tucol-kraeftig3,mark=none,line width=1] table[x index=0, y index=2, col sep=comma] 
		{data/plot_gw0005_phoconfnet_tcurve.csv};

		\addplot[tucol-rot2,mark=none,line width=1] table[x index=0, y index=3, col sep=comma] 
		{data/plot_gw0005_phoconfnet_tcurve.csv};

		\draw[black] (0,95) -| (5000,95); 

		\draw[black,dashed] (568,0) -| (568,600);
	\end{axis}

    	\node[above left=0.45cm and -0.8cm of gw_tcurve.east] (node_legend) {};
    	\begin{axis}[%
    	at={(node_legend.east)}, anchor=west,
    	hide axis,
    	width=0.1\textwidth,
    	height=0.1\textheight,
    	xmin=0, xmax=1,
    	ymin=0, ymax=1,
    	legend columns=1,
    	legend style={
    	    draw=white!15!black,
    	    legend cell align=left,
    	    /tikz/column 2/.style={
    	        column sep=15pt,
    	    },
    	    at={(1.2, 1.1)},
    	    anchor=north
    	},
	/pgfplots/legend image code/.code={	
		\draw[mark repeat=2,mark phase=2,#1] 
        	    plot coordinates {
                	(0cm,0cm) 
                	(0.1cm,0cm)
                	(0.2cm,0cm)
                	(0.3cm,0cm)
            	};
	}
    	]
    		\addlegendimage{tucol-kraeftig1!70,mark=none,line width=1pt}
		\addlegendentry{$\mapT$};

		\addlegendimage{black,mark=none,line width=1pt}
		\addlegendentry{$\text{mAP}_\text{ID}$};

    	    	\addlegendimage{tucol-kraeftig3,mark=none,line width=1pt}
		\addlegendentry{$\mrT$};
    		
    	    	\addlegendimage{tucol-rot2,mark=none,line width=1pt}
    		\addlegendentry{coverage};
    	\end{axis}
\end{tikzpicture}